\documentstyle[a4wide,named,twocolumn,11pt]{article}
\newcommand{\concat}{\mbox{$\cdot$}}
\begin{document}
\bibliographystyle{named}
\renewcommand{\dbltopfraction}{0.7}
\input{epsf}

\title{Best-First Surface Realization}
\author{Stephan Busemann\thanks{This work has been supported by a grant from 
The Federal Ministry for Research and Technology (FKZ~ITW~9402). I am
grateful to Michael Wein, who implemented the
interpreter, and to Jan Alexandersson
for influential work on a previous version of the system. Finally, I
wish to thank two anonymous reviewers for useful suggestions.
All errors contained in this paper are my own.}\\
DFKI GmbH\\
Stuhlsatzenhausweg 3\\
D-66123 Saarbr\"{u}cken\\
email: {\tt busemann@dfki.uni-sb.de}}
\date{}
\maketitle
\begin{abstract}
Current work in surface realization concentrates on the use
of general, abstract algorithms that interpret large, reversible grammars.
Only little attention has been paid so far to 
the many small and simple applications that require coverage of
a small sublanguage at different degrees of sophistication. 
The system TG/2 described in this paper
can be smoothly integrated with deep generation processes,
it integrates canned text, templates, and context-free rules into a
single formalism, it allows for both textual and tabular output, and
it can be parameterized according to linguistic preferences.
These features are based on suitably restricted production system
techniques and on a generic backtracking regime.
\end{abstract}

\section{Motivation} 
Current work in surface realization concentrates on the use
of general, abstract algorithms that interpret declaratively defined,
non-directional grammars.
It is claimed that this way, a grammar can be reused for parsing and
generation, or a generator can interpret different grammars (e.g.\
in machine translation). A prominent example for this type of
abstract algorithm is semantic-head-driven generation
\cite{Shi:Noo:Moo:90} 
that has been used with HPSG, CUG, DCG and several other formalisms.

In practice, this type of surface realization has several drawbacks.
First, many existing grammars have been developed with parsing as the
primary type of processing in mind. Adapting their semantics layer to a
generation algorithm, and thus achieving reversibility, 
can turn out to be a difficult enterprise
\cite{Rus:War:Car:90}. Second, many linguistically motivated 
grammars do not cover 
common means of information presentation, such as filling in a table,
bulletized lists, or semi-frozen formulae used for greetings in letters. 
Finally, the grammar-based 
logical form representation hardly serves as a suitable
interface to deep generation processes. 
Grammar-based semantics is, to a large extent, a compositional
reflex of the syntactic structure and hence corresponds too closely
to the surface form to be generated. As a consequence, only little
attention has been paid to interfacing this type of realizers 
adequately to deep generation
processes, e.g.\ by allowing the latter to influence the order
of results of the former.

The system TG/2, which is presented in this contribution,
overcomes many flaws of grammar-based surface realization systems
that arise in concrete applications. In particular, TG/2
\begin{itemize}
\item can be smoothly integrated with 'deep' generation processes,
\item integrates canned text, templates, and context-free rules into a
single formalism,
\item allows for both textual and tabular output,
\item efficiently reuses generated substrings for additional solutions, and
\item can be parameterized according to linguistic properties
(regarding style, grammar, fine-grained rhetorics etc.).
\end{itemize}

TG/2 is based on restricted production system techniques that
preserve modularity of processing and linguistic knowledge, hence
making the system transparent and reusable for various applications. Production
systems have been used both for modeling human thought (e.g.\ \cite{Newell:73})
and for the construction of knowledge-based expert systems (e.g.\
\cite{Shortliffe:76}). In spite of the modularity gained by separating
the rule basis from the interpreter, production systems have disappeared from
the focus of current research  because of their limited transparency
caused by various types of side effects. In particular, side effects 
could modify the data base in such a way that other rules become
applicable \cite{Dav:Kin:77}.

However, precondition-action pairs can be used in a more restricted way,
preserving transparency by disallowing side effects that affect the
database. In TG/2 preconditions are tests over the database contents
(the generator's input structure), and actions typically
lead to a new subset of
rules the applicability of which would be tested on some
selected portion of the database. By constraining the effects of
production rules in such a way, the disadvantages of early
production systems are avoided. At the same time, considerable
flexibility is maintained with regard to linguistic knowledge used.
A production rule may 
\begin{itemize}
\item involve a direct mapping to surface forms (canned
text), 
\item require to fill in some missing portion from a surface text
(template), or
\item induce the application of other rules (classical grammar rules)
\end{itemize}

Early template-based generation methods have correctly been criticized for
beeing too inflexible to account adequately for the communicative and 
rhetorical demands of many applications. On the other hand, templates
have been successfully used when these demands could be hard-wired
into the rules. In TG/2 the rule writer can choose her degree of
abstraction according to the task at hand. She can freely intermix
all kinds of rules.

The rest of the paper is organized as follows. TG/2 assumes as its input a
predicate-argument structure, but does not require any particular
format. Rather, a separate translation step is included that translates the
output of feeding components into expressions of 
the Generator Interface Language (GIL) (Section~\ref{GIL}). 
In Section~\ref{TGL} the formalism TGL (Template Generation Language)
for production rules is introduced. 
The properties of TGL allow for efficient generation of all
possible solutions in any order. The TGL interpreter and its
generic backtracking regime are presented in Section~\ref{btrk}.
It is used to parameterize
TG/2 by inducing an order in which the solutions are generated
(Section~\ref{para}). 

\begin{figure*}[t]
\epsfysize=14.5cm
\epsfbox{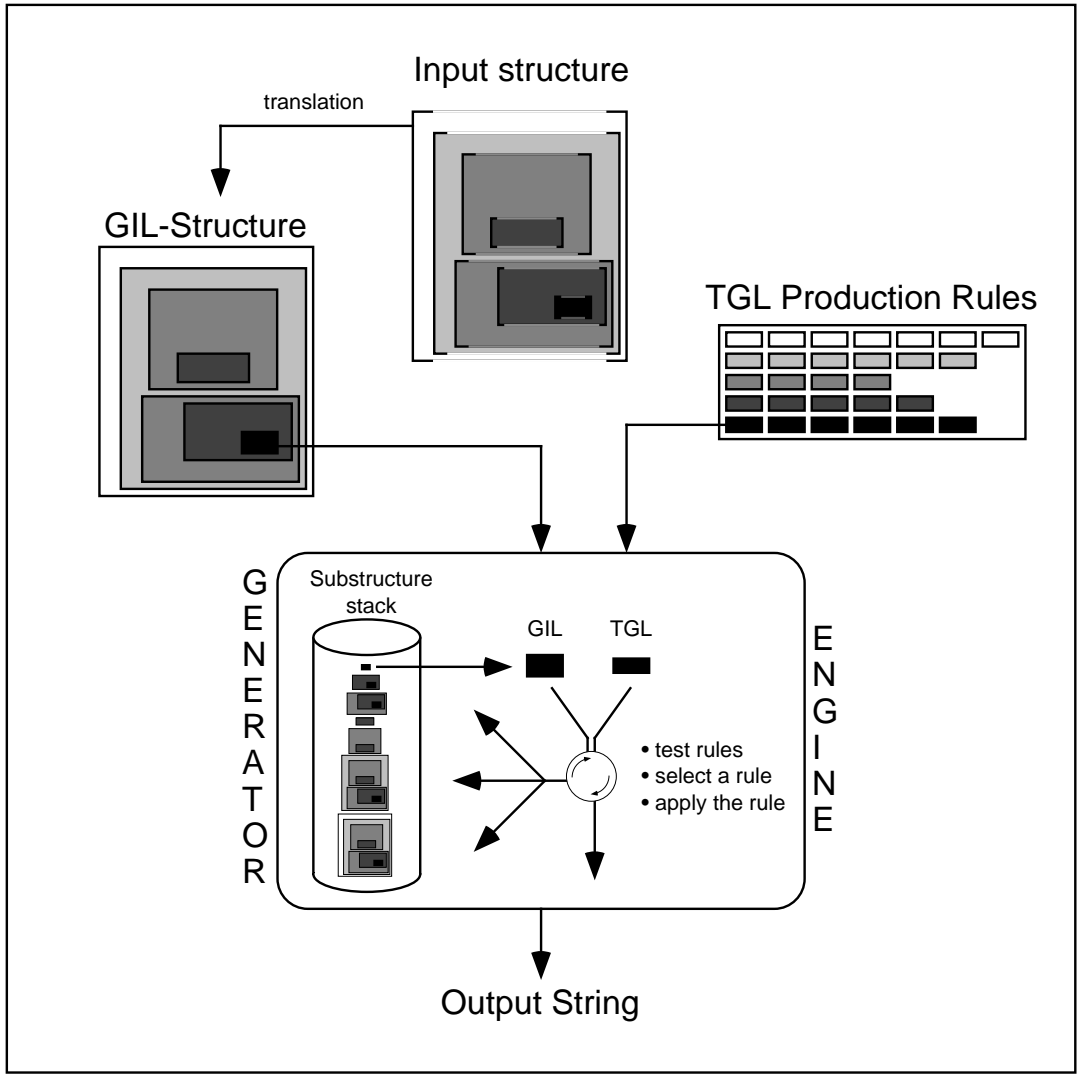}
\caption{Overview of the system TG/2.}
\label{figure}
\end{figure*}

Figure~\ref{figure} gives an overview of the
system and its components.

\section{The Generation Interface Language (GIL)} 
\label{GIL}
Although the level of
logical form is considered a good candidate for an interface to surface
realization, practice shows that notational idosyncrasies can pose
severe translation problems. TG/2 has an internal language, GIL, that
corresponds to an extended predicate argument structure. GIL is the
basis for the precondition test predicates and the selector functions
of TGL. Any input to TG/2 is first translated into GIL before being
processed. It is of considerable practical benefit to keep the
rule basis as independent as possible from external conditions (such
as changes to the output specification of the feeding system).

\begin{figure*}
\begin{center}
{\small
\begin{verbatim}
     [(PRED request)
      (HEARER [(ID refo365) (SET < nussbaum >)])
      (SPEAKER [(ID refo752) (SET < digisec >)])
      (THEME [(SMOOD [(TOPIC #1) (MODALITY unmarked) (TIME pres)])
              (PRED meet)
              (DREF [(ID refo610) (SET < meet1 >)])
              (ARGS < #1= [(ROLE agent)
                           (CARD single)
                           (CONTENT [(DREF [(ID refo621) (SET < zweig >)])
                                     (QFORCE noquant)
                                     (PRED humname)
                                     (NAME [(TITLE \"Prof.\")
                                            (SURNAME \"Zweig\")
                                            (SORT female)])])],
                      [(ROLE patient)
                       (CARD single)
                       (CONTENT [(DREF [(ID refo365) (SET < nussbaum >)])
                                 (QFORCE iota)
                                 (PRED object)])] >)
              (TIME-ADJ [(ROLE on) (CONTENT [(WEEKDAY 5)])])])]
\end{verbatim}
}
\end{center}
\caption{A sample GIL input structure ({\em Prof. Zweig will Sie am Freitag 
treffen\/} [Prof. Zweig wants to meet you on Friday]. {\tt <} and {\tt >} are
list delimiters; {\tt \#} denotes coreferences.}
\label{GIL-sample}
\end{figure*}

GIL is designed to be a target language suited for deep generation
processes. Similar aims have been pursued with the development of the 
Sentence Plan Language (SPL) \cite{Kas:Whi:89} that is used in a variety of
generation systems. Like SPL, GIL assumes only little grammatical
information. GIL can represent DAG-like feature
structures. 
Figure~\ref{GIL-sample} contains a sample GIL expression. The example shows
the major language elements:
\begin{itemize}

\item The top level consists of a speech act predicate and arguments
for author, addressee and theme (the speechact proper).
\item Discourse objects can be assigned unique constants ({\tt ID}) that denote
{\tt SET}s of discourse objects. 

\item {\tt SMOOD} expresses sentence modalities including sentence type, time,
a specification of
which constituents to topicalize in a German declarative sentence, etc.

\item The predicate argument structure is reflected by corresponding
features; {\tt ARGS} contains a list of arguments.

\item Different sorts of free temporal and local adjuncts can be
specified by corresponding features. In Figure~\ref{GIL-sample}, a
temporal adjunct is represented under {\tt TIME-ADJ}.

\item Arguments and, in part, adjuncts are specified for their role, for
cardinality, for quantificational force (under {\tt CONTENT.QFORCE}),
and further details such as name strings and natural gender.

\item Temporal adjuncts relate to some context (e.g. {\em tomorrow\/}) or
are indexical (e.g.\ {\em on Wednesday, February 7, 1996\/}). All
common combinations in German are covered.

\end{itemize}

\section{The Template Generation \protect\newline Language (TGL)} 
\label{TGL}

TGL defines a general format for expressing production rules as
precondition-action pairs (cf.\ Figure~\ref{syntax}). A TGL rule is
applicable if its preconditions are met. A TGL rule is successfully
applied, if the action part has been executed without failure.
Failure to apply a TGL rule signals that the
rule does not cover the portion of the input structure submitted to it.

\begin{figure*}
{\small
\begin{verbatim}
      <rule>      ::= (DEFPRODUCTION <string> <tgl-rule>)

      <tgl-rule>  ::= (:PRECOND (:CAT <category>
                                 :TEST (<lisp-code>+))
                       :ACTIONS (:TEMPLATE <template>+
                                 {:SIDE-EFFECTS <lisp-code>}
                                 {:CONSTRAINT <feature-equation>+}))

      <category>  ::= TXT | S | VP | NP | PP | PPdur | INF | ADJ | ...

      <template>  ::= (:RULE <category> <lisp-code>) |
                      (:OPTRULE <category> <lisp-code>) |
                      (:FUN <lisp-code>) |
                      <string>
\end{verbatim}
}
\caption{An excerpt of TGL Syntax.}
\label{syntax}
\end{figure*}
 
Figure~\ref{sample} shows a sample TGL rule. It corresponds to an
infinitival VP covering 
a direct object, an optional temporal adjunct, an optional expression
for a duration (such as {\em for an hour\/}), 
an optional local adjunct (such as {\em at the DFKI building\/}), and the
infinite verb form. 
Given the input GIL structure of Figure~\ref{GIL-sample}, the 
VP {\em Sie am Freitag treffen\/} [to meet you on Friday]
could be generated from this rule. Among the optional constituents,
only the temporal adjunct would find appropriate material in the
GIL input structure (under {\tt THEME.TIME-ADJ}). 

Every TGL rule has a unique name, denoted by the initial string. This
name is used for expressing preferences on alternative rules (cf.\
Section~\ref{para}).

\begin{description}
\item[Category:] The categories can be defined as in a context-free grammar.
Correspondingly, categories are used for rule selection (see below).
They ensure that a set of TGL rules possesses a context-free backbone.

\item[Test:]
The Lisp code under {\tt :TEST} is a boolean predicate
(usually about properties of the portion of input structure under
investigation or about the state of some memory). In the sample rule,
an argument is required that fills the patient role.

\item[Template:] 
Actions under {\tt :TEMPLATE}\footnote{The notion of template is preserved
for historical reasons: the predecessor system TG/1 was strictly
template-based.}  include the selection of other rules ({\tt
:RULE, :OPTRULE}), executing a function ({\tt :FUN}), or returning an
ASCII string as a (partial) result.

When selecting other rules by virtue of a category, a Lisp function is
called that identifies the
relevant portion of the input structure for which a candidate rule
must pass its associated tests. In Figure~\ref{sample}, the first action
selects all rules with category {\tt NP}; the
relevant substructure is the argument filling the patient role (cf.\
the second element of the {\tt ARGS} list in Figure~\ref{GIL-sample}).
If there is no such substructure, an error
is signalled\footnote{In the case at hand, the grammar writer preferred
to ensure availability of the substructure by virtue of the test
predicate.} unless an {\tt OPTRULE} slot (for ``optional rule'') was
executed. In this case, processing continues without results from that slot.

Functions  must return an ASCII string. They are mostly used
for word inflection; otherwise, for German every inflectional variant would have
to be encoded as a rule. TG/2 uses the morphological inflection component
MORPHIX \cite{Fin:Neu:88}.

\begin{figure*}
\begin{center}
{\small
\begin{verbatim}
       (defproduction "VPinf with temp/loc adjuncts"
               (:PRECOND (:CAT VP
                          :TEST ((role-filler-p 'patient)))
                :ACTIONS (:TEMPLATE (:RULE NP (role-filler 'patient))
                                    (:OPTRULE PP (temp-adjunct))
                                    (:OPTRULE PPdur (temp-duration))
                                    (:OPTRULE PP (loc-adjunct))
                                    (:RULE INF (theme))
                          :CONSTRAINTS (CASE (NP) :VAL 'akk))))
\end{verbatim}
}
\end{center}
\caption{A sample production rule for a VP with an infinitive verb form
placed at the end.}
\label{sample}
\end{figure*}
\item[Side effects:]  
The Lisp code under
{\tt :SIDE-EFFECTS} is a function whose value is ignored. 
It accounts for non-local dependencies between
substructures, such as  updates of a discourse memory. Note that
these effects can be traced and undone in the case of backtracking.

\item[Constraints:] 
Agreement relations are encoded into the rules
by virtue of a \mbox{PATR} 
style \cite{Shi:Usz:Per:83} feature percolation mechanism. The rules can
be annotated by equations that either assert equality of a feature's 
value at two or more constituents or introduce a feature value at
a constituent. Attempting to overwrite 
a feature specification yields an error. 
In Figure~\ref{sample}, the right-hand side constituent 
NP is assigned accusative case.  
Any of these effects are subject to backtracking.
\end{description}

Using TGL, small task- and domain-specific grammars can be written
quickly. For instance, in the domain of appointment scheduling the
system {\sc Cosma}  \cite{Bus:Oep:Hin:94}  has to accept, reject, 
modify, or refine suggested meeting dates via email. The sublanguage
encoded in TGL only needs a few speech acts, about twenty sentential
templates, and a complete account of German date expressions. Moreover,
formal as well as informal opening and closing phrases for emails are covered. 

Larger grammars may become difficult to maintain unless special care is
taken by the grammar writer to preserve a global structure of rules, 
both by defining suitable  categories and by documenting the rules. 
TGL rules are presently written using a text editor. A specialized
TGL grammar editor could improve the development and the organization
of grammars considerably.
Syntactic correctness is checked at
compile-time by an LR-Parser generated by Zebu \cite{Laubsch:92b}
on the basis of a BNF syntax for TGL.

\section{An interpreter with \protect\newline generic backtracking} 
\label{btrk}
TG/2 has a simple interpretation procedure that corresponds to
the classical three-step evaluation cycle in production systems
(matching, conflict resolution, firing) \cite{Dav:Kin:77}.
The algorithm receives a GIL structure as its input and uses
a distinguished category, {\tt TXT}, to start from.
\begin{description}
\item[1. Matching:] 
Select all rules carrying the current category.
Execute the tests for each of these rules on the input structure
and add those passing their test to the {\em conflict set}.
\item[2. Conflict resolution:] Select an element from the conflict set.
\item[3. Firing:] 
Execute its side effect code (if any). Evaluate its constraints (if any).
For each action part, read the category, determine the
substructure of the input by evaluating the associated function, and
goto 1.
\end{description}

The processing strategy is top-down and depth-first. The set
of actions is fired from left to right. Failure of executing some
action causes the rule to be backtracked. 

The interpreter yields all solutions the grammar can generate. It
attempts to generate and output a first solution, producing
possible alternatives only on external demand. Any alternative is based
on backtracking at least one rule. Backtrack points correspond to
conflict sets containing more than one element.
\begin{figure*}
\begin{center}
\begin{tabular}{|l|l|c|r|} \hline

          & \multicolumn{1}{c|}{\em pre context}
          & \multicolumn{1}{c|}{\em ego}
          & \multicolumn{1}{c|}{\em post context}                 \\ \hline

$B_1$     & $s_1$                                                 %
          & $V_1 = \{s_{2i} | 1 \leq i \leq |B_1|\}$              %
          & $s_3 \concat V_2 \concat s_8$                      \\ \hline

$B_2$     & $s_1 \concat V_1 \concat s_3$                         %
          & $V_2 = \{s_{4j} | 1 \leq j \leq |B_2|\}$              %
          & $s_8$                                                 \\ \hline
                                                                  \hline
$B_{2_1}$ & $s_1 \concat V_1 \concat s_3 \concat s_{5j}$         %
          & $V_{2_1} = \{s_{6k} | 1 \leq k \leq |B_{2_1}|\}$     %
          & $s_{7j} \concat s_8$                                 \\ 
        & & where $s_{4j} = s_{5j} \concat V_{2_1} \concat s_{7j}$ & \\ \hline
\end{tabular}
\end{center}
\caption{Table of Backtrack Points: $B_2$ is encountered outside of the
ego of $B_1$. $B_{2_1}$ is encountered inside the ego of $B_2$.}
\label{btrk-fig}
\end{figure*}

Backtracking may turn out to be inefficient if it involves
recomputation of previously generated substrings. In TG/2 this effort
is reduced considerably because it is only necessary to recompute
the part licensed by the newly selected rule. What has been
generated before or after it remains constant (modulo some word forms
that need to agree with new material) and can thus be
reused for subsequent solutions. This is possible
due to the design properties of TGL: rules cannot
irrevocably influence other parts of the solution. In particular, the
context-free backbone implicit in any solution and the restrictions
to side effects mentioned above keep the structural effects of TGL rules local.

In the sequel, technical aspects of the backtracking regime are
discussed. 
Let us assume that the interpreter compute a backtrack point. 
Let us call the sequence of strings 
generated by previous actions its {\em pre-context}, the set of
string sequences
generated from the elements of the conflict set its {\em ego}, 
and the sequence of strings generated 
from subsequent actions its {\em post-context}. For every ego, the pre- or 
the post context may be empty.

Each time a backtrack point is encountered during processing, 
an entry into a global table is made
by specifying its pre-context (which is already known due to the
left-to-right processing), a variable for the ego (which
will collect the sequences of strings generated by the elements
of the conflict set), 
and a variable for the post-context (which is unknown so far).\footnote{In 
fact, it is preterminal rather than terminal elements that are stored in
the table in order to account for modified
constraints. This can be neglected in the present 
discussion, but will be taken up again below.}
Figure~\ref{btrk-fig} shows the state of a sample table
comprising three backtrack points  after all solutions have been
computed. The ego variable is shown using indices running over
the elements of the respective conflict sets.  
The operator `\concat' denotes concatenation of strings with strings or 
sets of strings, delivering all possible combinations. 

After the first solution has been found (i.e. $s_1 \concat s_{21}
\concat s_3 \concat s_{51} \concat s_{61} \concat s_{71} \concat s_8$), 
every ego set contains one element. The post contexts for all backtrack
points can be entered into the table. 

The next solution is generated by selecting anyone of the backtrack
points and adding a new element to the ego set. At the
same time, all other entries of the table are updated, and the set of
additional solutions can be read off straightforwardly from the entry
of the backtrack point just processed.
Assume, for instance, that $B_{2_1}$ generates a second solution, thus
causing $V_{2_1}$ to have two elements. We then get $s_1 \concat s_{21}
\concat s_3 \concat s_{51} \concat s_{62} \concat s_{71} \concat s_8$.
Now assume that $B_1$ also generates a second solution. This directly
yields two more solutions since the post context of $B_1$ includes,
via $s_{4j}$, the two elements of $V_{2_1}$.

This way only  the alternative elements of a conflict set have to be 
expanded from scratch. All other material can be reused. This is highly 
efficient for backtrack points introducing ``cheap'' alternatives 
(e.g. different wordings). Since the ego must be recomputed from scratch,
much less is gained with backtrack points occurring at a higher level (e.g.\
active vs. passive sentence). In order to avoid having to recompute 
successfully generated partial results within the ego, such results are 
stored during processing together with the part of the input structure
and the current category. They can be reused when passing an applicability
test that requires the stored category and input structure to be identical
to the current ones.

The backtracking approach described is based on the assumption that 
any constraints introduced for some ego can be undone and recomputed on the
basis of rules generating an alternative ego. 
Clearly, features instantiated for some ego may have effects onto 
the pre- or post-context. If an agreement feature receives a different value
during backtracking and it relates to material outside the ego, 
inflectional processes for that material must be computed again. 
These cases can be detected by maintaining a trace of
all constraint actions. The recomputation is rendered possible by adding,
in addition to storing terminal strings in 
the table, the underlying calls to the inflection component as well.

\section{Parameterization} 
\label{para}
Parameterization of TG/2 is based on specifying the way how the
generic backtracking regime should operate. It can be
influenced with regard to
\begin{itemize}
\item the element in the conflict set to be processed next, and
\item the backtrack point to be processed next.
\end{itemize}
Both possibilities taken together allow a system that feeds TG/2 to
specify linguistic criteria of preferred solutions to be generated
first. 

The criteria are 
defined in terms of rule names, and a criterion is fulfilled
if some corresponding rule is successfully applied. We call such a
rule {\em c-rule}.
TG/2 implements a simple strategy that processes those backtrack points
first that have conflict sets 
containing c-rules, and preferrably choses a c-rule from a
conflict set. When applied incrementally, this procedure
yields all solutions fulfilling (some of) the criteria first. 

It would be desirable to see the solution fulfilling most criteria first.
However, incremental application enforces decisions to be taken locally
for each conflict set. Any c-rule chosen may be the last one in a
derivation, whereas chosing a non-c-rule may open up further
opportunities of chosing c-rules. 
These limits are due to a lack of look-ahead information: it is not known in 
general which decisions will have to be taken 
until all solutions have been generated.\footnote{Note that this
conclusion  does not depend on the processing strategy chosen.} 
Clearly, sacrificing incrementality is not what should be desired
although it may be acceptable for some
applications.  The drawbacks include a loss
of efficiency and run-time. This leaves us with two possible
directions that can lead to improved results.

{\bf Analyzing dependencies of criteria:}
The solution fulfilling most criteria is generated first if  
sets of mutually independent
criteria are applied: fulfilling one criterion must not exclude the
applicability of
another one, unless two criteria correspond to rules of the same conflict
set. In this case, they must allow for the the application of the
same subset of criteria. 
If these conditions are met, chosing a c-rule from every
conflict set, if possible, will lead to a globally best solution first.
There is, however, the practical problem that the
conditions on the criteria can only be fulfilled by analyzing, and
possibly modifying, the TGL grammar
used. This contradicts the idea of having the user specify her preferences
independent of TG/2 properties.

{\bf Learning dependencies of criteria:}
Missing look-ahead information could be acquired  automatically by
exploiting the derivational history of previously generated texts.
For every applied rule, 
the set of c-rules applied later in the current subtree of a derivation
is  stored.
From this information, we can derive off-line for any set of criteria
which c-rules have applied in the corpus
and how often each c-rule has applied within a
derivation. Computing such information from 
the context-free backbone of TGL grammars instead
would be less effective since it
neglects the drastic filtering effects of preconditions.
However, checking the grammar this way indicates which c-rules will {\em not\/}
appear in some subtree.

During processing, TG/2 can then judge the global impact of chosing 
the locally best c-rule and decide to fulfill or violate a criterion.
The success of this method depends on how well the derivation under
construction fits with the sample data. 
The more examples the system observes, the more reliable will be its
decisions. 

The latter approach is in fact independent on how the criteria influence each
other. In addition, it can be extended to cope with {\em weighted\/} criteria.
A weight is specified by the user (e.g.\ a feeding system) and expresses the
relative importance of the criterion being fulfilled in a solution.
TG/2 would give preference to derivations leading to the
maximum global weight. The global
weight of a solution is the sum 
of the c-rule weights, each divided by the number of times the c-rule occurs.

However, different GIL structures may, for a TGL rule, lead to
different sets of follow-up c-rules. This causes
the decision to be nondeterministic unless the reasons for the
difference are learned and applied to the case at hand. We must leave it to
future research to identify and apply suitable learning algorithms to
solving this problem. 

Criteria have been implemented for choosing a language,
for chosing between active and passive 
sentences, for preferring paratactical over hypotactical style, and for 
choice of formal vs.\ informal wordings. Additional uses could include
some rhetorical structuring (e.g.\ order of nucleus and satellites
in RST-based analyses \cite{Man:Tho:88}).

The approach presented offers a technical framework 
that allows a deep generation process 
to abstract away from many idiosyncrasies of
linguistic knowledge by virtue of meaningful weighting functions.
Ideally, these functions must implement a theory of how
mutual dependencies of criteria should be dealt with.
For instance, lexical choice and constituent order constraints 
may suggest the use of passive voice (cf.\ e.g.\ \cite{Danlos:87a}).
It is a yet open question whether such a theory can be encoded by
weights. However, for some sets of preferences, this approach
has proven to be sufficient and very useful.

\section{Conclusion} 

In this contribution, we have introduced TG/2, a production-rule based
surface generator that can be parameterized to generate the best solutions
first. The rules are encoded in TGL, a language that allows the definition of
canned text items, templates, and context-free rules within the same formalism.
TGL rules can, and should, be written with generation in mind, i.e.\ the goal
of reversibility of grammars pursued with many constraint-based approaches
has been sacrificed. This is justified because of the limited usefulness
of large reversible grammars for generation. 

TGL is particularly well suited for the description
of limited sublanguages  specific to the domains and
the tasks at hand. Partial reuse of such descriptions depends
on whether the grammar writer keeps general, reusable definitions 
independent from the specific, non-reusable parts of the grammar.
For instance, time and date descriptions encoded for the {\sc Cosma}
domain can be reused in other TG/2 applications. 
On the other hand, TGL sublanguage grammars can be developed using existing
resources. For instance, suitable fragments of context-free grammars 
translated into TGL could be augmented by the domain and task specific 
properties needed. Practical experience must show whether this approach
saves effort.

The system is fully implemented in Allegro Common Lisp and runs on
different platforms (SUN workstations, PC, Macintosh). Computing 
the first solution of average-length sentences (10--20 words) takes between
one and three seconds on a SUN SS~20.
TG/2 is being used in the domain of appointment scheduling within DFKI's
{\sc Cosma} system. In the near future, the system will be used within an 
NL-based information kiosk, where information about environmental data must be
provided in both German and French language, including tabular presentations if
measurements of several substances are involved. 

\bibliography{cl-general-cross,cl-pub-cross,cl-general,cl-pub,update}
\end{document}